# Heuristic Optimization of Electrical Energy Systems: Refined Metrics to Compare the Solutions

Gianfranco Chicco, Andrea Mazza*

Politecnico di Torino, Dipartimento Energia "Galileo Ferraris", corso Duca degli Abruzzi 24, 10129 Torino, Italy

(e-mail: gianfranco.chicco@polito.it, andrea.mazza@polito.it)

* Corresponding author: A. Mazza (andrea.mazza@polito.it)

**Abstract**

Many optimization problems admit a number of local optima, among which there is the global optimum. For these problems, various heuristic optimization methods have been proposed. Comparing the results of these solvers requires the definition of suitable metrics. In the electrical energy systems literature, simple metrics such as best value obtained, the mean value, the median or the standard deviation of the solutions are still used. However, the comparisons carried out with these metrics are rather weak, and on these bases a somehow uncontrolled proliferation of heuristic solvers is taking place. This paper addresses the overall issue of understanding the reasons of this proliferation, showing a conceptual scheme that indicates how the assessment of the best solver may result in the unlimited formulation of new solvers. Moreover, this paper shows how the use of more refined metrics defined to compare the optimization result, associated with the definition of appropriate benchmarks, may make the comparisons among the solvers more robust. The proposed metrics are based on the concept of first-order stochastic dominance and are defined for the cases in which: (i) the globally optimal solution can be found (for testing purposes); and (ii) the number of possible solutions is so large that practically it cannot be guaranteed that the global optimum has been found. Illustrative examples are provided for a typical problem in the electrical energy systems area – distribution network reconfiguration. The conceptual results obtained are generally valid to compare the results of other optimization problems.

**1. Introduction**

In many optimization problems, the solution space contains several local optima. The global optimum can be calculated either through exhaustive search over all the feasible solutions (for small-scale problems), or with a search over a limited number of solutions (when it can be guaranteed that the non-searched solutions cannot provide better results). Otherwise (e.g., for metaheuristic optimization methods based on random number extractions), it cannot be said that the global optimum has been reached.

The optimization procedure based on multiple executions of a heuristic method leads to obtain pseudo-optimal or best-so-far solutions in different ways:

- by running a given method with different parameters, also aiming at performing sensitivity analysis to determine the best values of the parameters;
- by running different methods (or variants of the same method), each one with a given set of parameters;
- by changing the seed for random numbers extraction (even by using the same parameters).

In the sequel, the term *solver* is generally used to indicate the execution of an optimization method or variant with a specific structure and set of parameters.

For a given optimization problem, the determination of the best solver[1] is one of the topics widely addressed in the literature. For this purpose, suitable metrics are needed in order to carry out effective comparisons among the methods or variants.

The Evolutionary Computation community is addressing the comparison among different solvers in a wide way, by considering a predefined set of problems to be solved, and defining dedicated metrics. For example, the *performance ratio* (defined by dividing the computation time of the solver by the minimum computation time obtained from all the solvers) is used to build the *performance profile* of the solver [1], which is the cumulative distribution function (CDF) of the performance ratio. For optimization problems with particularly high computational burden, the *data profile* [2] is constructed by using the computation time, the number of function evaluations, and a user-defined target on the value of the objective function. Both approaches have as main limitations the selection of the set of problems to analyze, because no general criteria have been defined yet.

---

[1] The inappropriateness of the concept "determining the best solver" is discussed in Section 3.

Other types of comparisons consider the outcomes of two algorithms run on the same problem. These include non-parametric statistical hypothesis tests (e.g., the unpaired Wilcoxon rank-sum test), which calculate the confidence interval around the mean (or the median) of the solutions [3]. However, focusing on the mean or median may be limitative, as the details on the specific location of the solutions are not included.

A number of scientific contributions in the electrical energy systems domain use simple indicators (such as best and worst solutions, mean value, standard deviation and median of the solutions). However, none of these metrics provides significant outcomes. In fact:

- The best (globally optimal) solution could be obtained by chance at any time from any solver, regardless of the overall set of solutions reached; the best solution could be even found (without being aware of it) in one of the solutions indicated by the operator to form the initial population used by the heuristic method, namely, before running the procedure. In this case, running the procedure would provide the best result immediately. However, would anybody conclude that the heuristic is perfect, or it was just a lucky initialization?
- The mean value of the solutions could be the same for solvers that provide very different solutions (some of which solutions could be close to the global optimum), or for solvers leading to many solutions concentrated around the mean value (in this case the best solution could be far from the global optimum). In this case, how to decide which solver is better?
- Using the standard deviation in addition to the mean value seems to improve the situation, but still with insufficient indications on the statistical distribution of the solutions.

In order to obtain more robust statistical information on the quality of the solutions, information from further probabilistic moments (e.g., skewness, kurtosis, etc.) could be used. However, the interpretation of the contribution of these probabilistic moments to establish whether the results obtained from one solver are better than the results of another solver is not straightforward.

In practice, mean and median are still used to rank the algorithms also in International competitions based on heuristic optimizations. For example, in the series of conferences on Evolutionary Computation (CEC), the criteria to rank the algorithms for constrained real parameter optimizations executed on one problem with a predefined number of runs are based on mean and median [4].

Specific competitions on heuristic optimization exist also in the power and energy systems area. For example, the 2017 IEEE competition on modern heuristic optimizers for smart grid operation [5] constructed two testbeds run for a given number of scenarios. Each scenario required the execution of a given number $N_{MC}$ of Monte Carlo objective function assessments, resulting in $N_{MC}$ best solutions. The score of each scenario was then calculated as the mean value of the $N_{MC}$ best solutions, and the total score was determined by summing up the scores of each scenario. Again, the best solution and the mean values were used to establish the scores.

This paper introduces a different rationale, aiming to construct performance indicators simple to be calculated and interpreted, and statistically more effective than the best value or the first probabilistic moments to represent the nature of the solutions.

The main contributions of this paper are:

a) An overview on some power and energy system problems addressed with heuristic methods, to show that several papers used different heuristic methods, variants and/or hybrid versions with the focus on testing a new method, without providing advances or insights on the problem itself.

b) A wide discussion on why the use of simple metrics to compare the performance of different optimization algorithms may lead to an undue and somehow uncontrolled proliferation of heuristic solvers. A conceptual scheme is introduced to represent the unsolvable nature of the problem of limiting the introduction of "new best solvers".

c) The formulation of more refined metrics to compare the optimization results in a statistically significant way for problems in which finding the global optimum cannot be guaranteed (extending the previous findings of the authors presented in [6]) and, as a new addition to the same framework, for problems with computable global optimum.

In particular, more refined metrics are defined to compare the optimization results in two specific cases:

- *Case G* (Global) – The globally optimal solution can be calculated in an acceptable computation time. Conceptually, there would be no need for running a new solver on a problem with known global optimum. However, testing a solver on test systems with known globally optimal solution may also be useful for a preliminary assessment of that solver before applying it to large-scale problems. This is for example the case of using predefined *benchmark functions* to test the solvers [4][7]. However, the classical benchmark functions are rather different with respect to the function used for applications on electrical systems. In fact, the electrical network introduces challenging aspects, such as *discrete* problem formulations and especially constraints difficult to be handled (e.g., the equality constraints given by active and reactive power flow balances, and the radiality of the network structures [8][9]).
- *Case R* (Relative) – The *number* of feasible solutions is so large that the global optimum cannot be found in a reasonable computation time. In this case, the pseudo-optimal solution can be identified only in a relative way as the best solution found so far. This situation occurs frequently in many problems due to of the combinatorial explosion of the number of possible solutions for large-scale systems.

The metrics introduced in this paper are based on the calculation of a number of solutions from different solvers, on the construction of the CDF of these solutions, and on the definition of specific indicators with intuitive geometric meaning. These indicators are based on the CDF of the solutions obtained from each solver and on a *reference CDF* constructed by considering the global optimum for *Case G* or a selected set of solutions for *Case R*.

The next sections of this paper are organized as follows. Section 2 recalls some typical power and energy system problems addressed by using heuristic methods. Section 3 discusses the issue of finding out the best solver, casting the problem into a perpetual motion conceptual scheme. Section 4 illustrates the use of refined performance metrics based on first-order stochastic dominance concepts. Section 5 introduces the specific problem (optimal distribution system reconfiguration) addressed in the illustrative examples. Section 6 shows the results of specific case study applications. The last section contains the concluding remarks.

**2. Typical power and energy problems solved with heuristic methods**

Various optimization problems in the power and energy systems area have been solved by using heuristic methods, in particular with:

- objective functions defined in highly non-linear forms, with several local optima;
- linear or non-linear constraints that make the domain of definition of the variables mathematically non-convex;
- problems with no possibility to compute derivatives that could be sent to algorithms that use this type of information;
- discrete combinatorial problems, affected by the curse of dimensionality when applied to large-scale systems.

Some favorable aspects of using heuristic methods are that:

- the heuristics are generally simple to implement, even though in some cases careful implementation is needed to guarantee the effective incorporation of equality and inequality constraints (e.g., the radiality of the distribution network) [8];
- the heuristics contain in their underlying principles some specialized operators (e.g., parallelism, elitism, selection with probability-based acceptance, topology, memory, immunity, and self-adaptation [10]) that could enable wider exploration of the space of the solutions;
- the heuristics can be hybridized to benefit from the different characteristics of two or more methods, for example adding an effective local search strategy to a method based on exploring a wide range of solutions, or using an internal optimization to find the most suitable parameters for the solver. A successful example is the Evolutionary Particle Swarm Optimization (EPSO) [11], which uses an evolutionary model together with a particle movement operator to formulate a self-adaptive algorithm.

Various applications of heuristic methods to electrical power and energy systems have been reviewed in [12]. A selected set of illustrative applications is considered in this paper to show the diffusion of heuristic methods applied to these problems, also in recent years:

- *Distribution system reconfiguration* (DSR): selection of the open/closed condition of the distribution network branches (or of the switches at the branch terminals) to optimize a predefined objective (or multi-objective) function. DSR is a combinatorial optimization problem that uses non-linear mathematical models. Specific constraints are the network connectivity, the need of considering only radial configurations, and the technical limits on the node voltages, branch currents, and others [10]. More details are indicated in Section 5.2, as the DSR problem is used in this paper to show the application examples.

- *Economic dispatch* (ED): calculation of the electrical power outputs of the generation units in the power system to obtain the minimum total operating cost, subject to a set of constraints. The equality constraints are the power balance equations. The inequality constraints (depending on the type of generation units) can include minimum and maximum power, prohibited operating zones, multiple fuel options, transmission limits, the valve-point effect, ramp rates, discharge limits, reservoir volume limits, and water dynamic balance equation with transport delay time [13].

- *Load forecasting* (LF): forecast of the demand at different time scales (short-term, medium-term, and long-term) and levels of aggregation (individual or aggregate loads). The demand depends on exogenous factors (e.g., temperature, humidity, economic and social aspects) in a complex and non-linear way, and other random components appear as "noise".

- *Maintenance scheduling* (MS): definition of the sequence in time to perform maintenance of the generation units to optimize a given objective function under the system operational constraints. This problem is non-linear and stochastic, and may have a number of conflicting objectives. MS is closely linked to reliability and costs, which can be handled within a risk analysis framework. The constraints are technical (e.g., load demand, system reserve requirements, availability of equipment, and duration of the maintenance period), personnel-related (e.g., availability of maintenance teams), and economic (budget constraints).

- *Optimal power flow* (OPF): determination of the operating point at steady state to minimize a predefined objective (or multi-objective) function (e.g., considering costs, emissions, system losses, and reactive power support) under a wide set of network constraints.

- *Power system planning* (PSP): the related problems include *operational* planning (i.e., changes occurring in the system at constant load, such as enhanced system automation, replacement of network components, or reactive power planning) and *expansion* planning of network and generation (considering scenarios of demand growth, also with the evolution of distributed generation and storage). The objective function for planning is typically economic and contains discounted investment costs, operation and maintenance costs, reliability costs, and salvage value. The constraints are numerous and involve technical, economic, environmental and social aspects. Planning problems are by nature large-scale, non-linear combinatorial problems, with integer and real variables. The number of solutions could grow exponentially with the system size, and there is typically a large number of local optima.

Among the contents presented in [12] in 2008, the heuristic methods surveyed in [14] include evolutionary algorithms (e.g., genetic algorithms, evolution strategies, evolutionary programming, and genetic programming), simulated annealing, tabu search (which is not a probability-based method), and particle swarm optimization. However, many other heuristic techniques have been introduced in the last decade. A text search has been carried out in the Science Direct database, considering the set of problems indicated above and different heuristics. Table 1 shows the number of articles in which each heuristic has been mentioned or applied (in its original version or in a variant), or has been used as a benchmark to compare the results obtained by another solver. The cases in which the heuristic appears for at least ten times for a given problem are listed in the table. Other heuristics appearing less than ten times are not considered. A basic reference is included for each heuristic.

From Table 1 it appears clearly that some classical heuristics are widely considered, but also many heuristics appeared recently have been applied. Overall, hundreds of articles are available in the literature to solve the set of problems indicated. For this reason, a comprehensive review of the contents of these contributions is outside the scope of this paper. However, most of these articles do not provide any advance or insight on the specific problem, but merely aim at testing a "new" heuristic, or a variant of an existing heuristic, or a hybridization of different methods (e.g., different heuristics, or heuristics and conventional algorithms, in the latter case typically calling the conventional algorithms to perform a deterministic local search). In the most significant cases, some parts of the methods are customized to deal with specific aspects of the problem analyzed (e.g., handling the radial network constraint in DSR). This customization can be seen as a methodological insight. Nevertheless, there is no certainty that the changes introduced lead systematically to better results with respect to other variants proposed. This lack of certainty gives room to the proliferation of new methods, variants and hybridizations.

Another reason why the number of contributions on testing heuristics in the power and energy systems area is so high is the lack of widely accepted benchmarks defined for specific problems. This situation is different with respect to the evolutionary computation community area, where given benchmark functions are used to test the algorithms. However, these benchmark functions typically have different properties with respect to the objective functions and constraints encountered in power and energy system problems. In particular, the major challenges appearing in the power and energy systems area include the need for handling the non-linear equality constraints that describe the power balances in large-scale applications, the topological constraints of the electrical networks, and the non-connected domain of definition of the decision variables.

On the basis of the above considerations, the next section presents a consistent scheme to explain the main reasons why there are so many heuristics proposed to solve the optimization problems, claiming each time that the solver tested is better than the existing ones.

## 3. Best Solver or Best Solution? A Perpetual Motion Conceptual Scheme

Let us consider an optimization problem that admits several local optima, among which the global optimum cannot be identified in a reasonable computation time. This problem is suitable for testing new solvers. Let us start from the best solution found so far, and construct a framework to represent the activity carried out by testing different solvers.

Fig. 1 shows how the identification of the best solver can be cast into a *perpetual motion* conceptual scheme, with no exit point. The solutions (i.e., objective function values) are determined in the *testing* block. On the basis of this scheme, the number of scientific publications that appear on the same subject by changing the solver is continuously increasing. Once a new "best" solver has been identified, the process continues by testing new solvers. Due to the intractable number of possible solutions, and to the uncertainty given by the random search embedded in the solvers, it is possible to find out another solver providing better results according with the metric established for comparison. Of course, if the metric is *weak*, better solutions may be found more easily, as remarked below.

An interesting question may be: is this growth in the number of publications continuously claiming to have found the best solver conceptually sound? The notes provided in this section aim at showing that in various cases the reasons to declare a "new" solver as the best one are somehow questionable.

The key issue is that, in the testing of a given problem with different solvers that provide stochastic outcomes, the relation between the solvers is *non-transitive*. For example, given three solvers denoted as A, B and C, the fact that solver A is better than solver B in a test, and that solver B is better than solver C in another test, does not imply that solver A is better than solver C in a further test. Thereby, even though a new solver has been tested on a given problem against other solvers, it cannot be concluded that the new solver is generally better than the other solvers. The main problem is that this approach is only driven by the goal of obtaining better results, without trying to understand *why* a solver performs better than others. This approach is making the meta-heuristics field vulnerable, as the undue and somehow uncontrolled proliferation of meta-heuristic solvers makes it difficult to recognize the true innovation occurring in the field [50].

In the framework indicated in Fig. 1 for *Case R* optimization problems with no known global optimum applied to large-scale systems, the crucial point is: *if the solutions found on a specific problem by using one solver are better than with another solver, this does not mean that the solver is better*. In other terms, "the solver is better" is a *global* property that cannot be demonstrated in a complete way. Thereby, any contribution that claims to have found the "best method" by comparing solutions obtained from different methods should at least restate the conclusions to indicate that the method tested has provided better solutions than other methods for the specific problem and case study application (with the specific initialization and in terms of the indicator used to assess the solutions). Of course, in this case the strength of the scientific contribution would sound poorer, almost vanishing.

Nevertheless, in the absence of a global optimum there is no way to break the perpetual motion conceptual scheme in a scientifically rigorous way. What can be done is to *improve the mechanism of comparison among different solvers*, requiring more significant information to declare that the solutions obtained from a solver are better than the solutions given by other solvers. In this case, the core of the problem is to reinforce the meaning of the test block "*better solution*?" appearing in Fig. 1. This can be done by resorting to refined metrics for ranking the solvers, based on a statistically significant set of solutions.

A dedicated metric has been introduced in [51], using two criteria to evaluate an algorithm, namely, the number of objective function calculations occurred before satisfying the stop criterion, and the value of the best objective function found by the optimization algorithm. These criteria are used in a Pareto-dominance analysis, concluding that the best algorithms are the ones that produce non-dominated solutions in the plane defined by the number of objective function calculations and by the objective function value itself. An approximation to the first-order stochastic dominance concept [52] has been used in [51] to rank the solvers in each evaluation criterion. Further studies in [53] considered also one-way ANOVA for ranking each evaluation criterion.

In the following section, the stochastic dominance concepts are exploited to formulate meaningful indicators for creating stronger comparisons among the optimization solutions. These indicators are then tested on the DSR problem, whose characteristics are more elaborated than the classical benchmark functions.

## 4. Refined Metrics and Indicators to Characterize the Results of Optimization Methods

*4.1. Optimization formulation and deterministic dominance*

The general framework for optimization considers the minimization case (without loss of generality) for a system with given vectors $p$ of parameters and $d$ of decision variables, and a scalar objective function $y$. In mathematical terms:

$$y = \min\{g(p, d)\} \qquad (1)$$

s.t. $r(p, d) = 0$

$\qquad w(p, d) \leq 0$

where $g(.)$, $r(.)$ and $w(.)$ indicate the function representing the objective, the vector of equality constraints, and the vector of inequality constraints, respectively.

The solvers are taken from a given set $S$. Each solver is executed $H_s$ times. The dependence on $s$ appears because the number of solutions may be different from solver to solver (for example, considering the execution times of each solver, $H_s$ can be determined as the number of solutions provided by the solver $s$ within a predefined total execution time).

The metrics considered here are based on the concept of *dominance* between two sets of solutions. Let us take the $H_A$ and $H_B$ solutions respectively provided by two solvers $\{A, B\} \in S$ run on the same optimization problem. If the solutions obtained from the two solvers are not overlapped, *deterministic dominance* does exist. Let $y_A^h$ be the $h^{th}$ solution obtained with solver A, and $y_B^h$ the $h^{th}$ solution obtained with solver B. The solutions obtained from solver A exhibit deterministic dominance over the solutions obtained from solver B if (Fig. 2):

$$\max_{h=1,\dots,H}\{y_A^h\} < \min_{h=1,\dots,H}\{y_B^h\} \qquad (2)$$

However, in practical applications, deterministic dominance rarely occurs. Further concepts of stochastic dominance are illustrated in the next section.

*4.2. Stochastic dominance*

The stochastic dominance concepts introduced in [52] are revisited and applied to the solutions of an optimization problem with different solvers, extending the approach recently introduced by the authors in [6].

Let us consider the same number $H$ of solutions for each solver to be compared. Let us construct the CDF $F_s^{(H)}(y)$ of the $H$ solutions of the optimization (1) obtained from each solver $s \in \mathbf{S}$, sorted in the ascending order of the variable $y$. Fig. 3 shows an example of construction of the CDF for an example with $H = 50$ solutions and objective function expressed in per units (p.u.).

Different stochastic dominance formulations can be introduced by considering the solvers $\{A, B\} \in \mathbf{S}$, run $H$ times each on the same minimization problem:

1. *First-order* stochastic dominance: the solutions obtained from solver A exhibit first-order stochastic dominance over the solutions obtained from solver B if and only if the following condition is satisfied on the CDFs for any value of the variable $y$, with strict inequality existing for at least one value of $y$:

$$F_A^{(H)}(y) \geq F_B^{(H)}(y) \tag{3}$$

   In graphical terms (Fig. 4a), the condition (3) means that no entry of the CDF referring to the solutions of solver B has to lie on the left-hand side of the CDF referring to the solutions of solver A. The strict inequality is required to exclude that all the entries in the two CDFs are identical.

2. *Second-order* stochastic dominance: the solutions obtained from solver A exhibit second-order stochastic dominance over the solutions obtained from solver B if and only if the following condition is verified on the CDFs for any value of the variable $y$, with strict inequality existing for at least one value of $y$:

$$\int_{z=0}^{y} \left( F_A^{(H)}(z) - F_B^{(H)}(z) \right) \geq 0 \tag{4}$$

   In graphical terms (Fig. 4b and Fig. 4c), the second-order stochastic dominance condition is based on an integral quantity. Thus, it cannot be excluded that one or more entries of the CDF referring to the solutions of solver B are located on the left-hand side of the CDF referring to the solutions of solver A (as in Fig. 4b). The case of Fig. 4b satisfies the second-order stochastic dominance conditions (as the area in Fig. 4c is always positive). However, it does not satisfy the first order stochastic dominance, because there are intersections between the CDFs.

By definition, the first-order stochastic dominance condition is sufficient to ensure second-order stochastic dominance. However, the reverse case does not hold, as shown before. For the same reason, higher-order stochastic dominance is not considered. The first-order stochastic dominance is then taken as the relevant metric for constructing the indicators shown in the next sections.

*4.3. Determination of the reference CDF*

The proposed performance indicators require the definition of a *reference CDF* $F_{ref}^{(H)}(y)$ constructed by considering $H$ solutions for the variable $y$. The entries of $F_{ref}^{(H)}(y)$ must satisfy the following property, with respect to the entries of every CDF $F_s^{(H)}(y)$ corresponding to the solvers $s \in \boldsymbol{S}$ used in the comparison:

$$F_{ref}^{(H)}(y) \geq F_s^{(H)}(y) \tag{5}$$

Practically, no entry of the CDFs originated from the solutions of the solvers under comparison has to be located on the left-hand side of the reference CDF. However, partial or even total superposition of the CDFs under test with the reference CDF is allowed. Thereby, the reference CDF exhibits first-order stochastic dominance with respect to all the CDFs under test, with the additional possibility that the reference CDF is exactly equal with one of the CDFs (not included in the first-order stochastic dominance conditions).

The reference CDF is constructed in a different way in the following two cases:

- *Case G*: the reference CDF $F_{ref,G}$ is equal to zero until the global optimum is reached, then it jumps to unity (Fig. 5).
- *Case R*: the procedure introduced in [6] is used. The reference CDF $F_{ref,R}^{(H)}$ is constructed by calculating the objective function for $H_s$ solutions from each solver, by using the penalized objective function to take into account possible constraint violations, as indicated in Section 5.2. Then, the number of solutions is taken as $H = \min_s \{H_s\}$, and the best $H$ solutions are considered for each solver, with a benefit for solvers that provide faster executions. To compare the solutions obtained from $M$ solvers, the number of solution points available is $M \cdot H$, from which the best $H$ points (taken in the ascending order) are used to construct the reference CDF. Fig. 6 summarizes the procedure by using three solvers $\{A, B, C\} \in \boldsymbol{S}$ and $H = 100$ points for each solver.

There is a conceptual difference between the rationale used to define the reference CDF for *Case G* and *Case R*. In *Case G* a fixed entry (the global optimum) is available, thereby the reference CDF is fixed and the comparison between the actual CDF and the reference CDF is conducted in *absolute* terms. In *Case R* the entries of the reference CDF may vary depending on $H$ and on the context in which the calculations are carried out (e.g., the computation time limit indicated above), so that the comparison between the actual CDF and the reference CDF is conducted in *relative* terms.

*4.4. Definition of the performance indicators*

For the calculation of the performance indicators, the reference CDF and the test CDF must be represented with the same number of points $H$ on the vertical axis, corresponding to the width $\Delta c = 1/H$ of each step on the vertical axis. Thereby, in *Case G* the reference CDF is conceptually discretized into $H$ points having the same value (the global optimum), located at successive vertical steps of width $\Delta c$.

The rationale to construct the performance indicators by using stochastic dominance concepts is that these indicators provide a *global* statistically-based information on the set of solutions considered. This kind of information describes the performance of each solver better than any individual statistical indicator (e.g., mean, median, standard deviation, …). Moreover, in the proposed approach the performance indicator is quantified by using a single number variable from 0 to 1, with higher values representing better performance, avoiding the need of merging the information from multiple statistical indicators to characterize the results.

At first, the *area* between the reference CDF and the test CDF is calculated. The $H$ objective function values located on the horizontal axis of the CDF, $z = 1,…, H$, are denoted with $y_s^H(z)$ for the test CDF of solver $s \in \mathbf{S}$, and with $y_{\text{ref,R}}^H(z)$ for the reference CDF in *Case R*. In *Case G*, there is a single value $y_G$. The following areas are calculated by summing up the horizontal areas defined from the reference and test CDFs (since the vertical steps in the CDF have the same width, Fig. 7):

- In *Case G*:

$$A_{s,G}^{(H)} = \frac{1}{H}\int_{z=0}^{H}\left(y_s^{(H)}(z) - y_G\right) \qquad (6)$$

- In *Case R*:

$$A_{s,R}^{(H)} = \frac{1}{H}\int_{z=0}^{H}\left(y_s^{(H)}(z) - y_{\text{ref,R}}^{(H)}(z)\right) \qquad (7)$$

The performance indicators are then defined under the general name *OPISD* (Optimization Performance Indicator based on Stochastic Dominance). For the two cases described in Section 4.3, the indicators are formulated as:

- *OPISD*$_G$ (*global OPISD*), applied in *Case G* by considering $H$ CDF values:

$$OPISD_G^{(H)} = \frac{1}{1+A_{s,G}^{(H)}} \qquad (8)$$

- *OPISD*$_R$ (*relative OPISD*), applied in *Case R* by considering $H$ CDF values:

$$OPISD_R^{(H)} = \frac{1}{1+A_{s,R}^{(H)}} \qquad (9)$$

The OPISD definition encompasses cases with null area (the CDF of the test solver is exactly equal to the reference CDF, with OPISD = *1*), or when the objective function assumes both positive and negative values, providing results in the interval (0,1) at all times.

In spite of their similar definition, the two indicators have a fundamental conceptual difference: the reference CDF for *global OPISD* is based on the global optimum, which never changes and is an absolute reference to rank the results of solvers executed in different situations. Conversely, the reference CDF for *relative OPISD* generally changes each time a comparison among two or more solvers is carried out, so that the numerical results for *OPISD*$_R$ are only valid for the corresponding ranking obtained, and cannot be used in successive comparisons.

The test methods compared are then *ranked* on the basis of the relevant *OPISD* indicator, from the most effective one (with the highest *OPISD* value) to the less effective one (with the lowest *OPISD* value).

The new *OPISD* indicators are a general-purpose way to rank the performance of *heuristic* methods, for which many solutions may be available from multiple executions (even with the same parameters, by just changing the seed for random number extractions). Applications of the proposed metric to the discrete optimization problem DSR are illustrated in Section 6.

*4.5. Further indicators*

In addition to simple indicators such as the mean, the median and other probabilistic moments, further indicators can be defined by taking into account how many solutions are located on the reference CDF. In this case, the different nature of *Case G* and *Case R* emerges again. In fact:

- The indicator $PERC_\text{G}^{(H)}$ represents the percentage of solutions that reach the global optimum in *Case G*. This indicator has an absolute meaning, as the global optimum for the problem under analysis is fixed.

- The indicator $PERC_\text{R}^{(H)}$ represents the percentage of solutions located on the reference CDF in *Case R*. This indicator has relative meaning, as the reference CDF in this case changes depending on the test CDFs in the comparison carried out.

## 5. Optimization Problem and Solution Methods

*5.1. Stochastic dominance applications to power and energy systems*

The most typical application of stochastic dominance concepts in the power and energy systems area refers to risk-aversion modelling in electricity trading and portfolio optimization. The second-order stochastic dominance constraint has been imposed in [54] to reduce the risk of low profit for an electricity retailer, and in [55] for self-scheduling large consumers. In [56] portfolio optimization has been carried out by setting up a minimum tolerable CDF used as the reference; then, the CDF of the portfolio model is accepted if it exhibits second-order stochastic dominance over the reference CDF.

Furthermore, first-order stochastic dominance constraints have been applied to optimize the introduction of new generation capacity [57], and in risk-based energy trading in a virtual power plant [58]. Energy scheduling in a residential system with renewable energy sources has been addressed in [59] by using first-order stochastic dominance to choose the renewable energy distributions. The first-order stochastic dominance used in [6] adopted a performance indicator corresponding to $OPISD_R$.

Without loss of generality, the results are presented on a classical optimization problem in the electricity distribution area – DSR with minimum losses. This discrete optimization problem is conceptually simple and can be easily understood by the readers with different backgrounds. At the same time, for small-scale systems is possible to calculate the global optimum and construct the reference CDF for *Case G*. Conversely, for large-scale systems the combinatorial explosion of the number of solutions prevents determining the global optimum in a reasonable computation time, so that the definitions valid for *Case R* have to be applied.

*5.2. Description of the DSR optimization problem*

The Medium Voltage (MV) electricity distribution systems are structurally composed of a number of nodes interconnected by branches forming a weakly meshed structure. The network is typically operated with radial configuration in order to simplify the protection schemes.

For a distribution system with $N$ nodes, $B$ total branches and $S$ supply points, any radial configuration starting from the supply nodes without isolating any other node has $\Lambda = B-N+S$ open branches. This condition is necessary to define the radial network, but it is not sufficient, because cutting $\Lambda$ branches at random would very likely create loops in the network and isolate some nodes. Moreover, the total number $\Psi$ of radial configurations that can be obtained by starting from a given network structure is computable from the Kirchhoff's tree matrix theorem [60]. However, the number $\Psi$ could be so high to make the generation of the configurations (with a graph-search algorithm) practically intractable. For example, the real distribution network shown in [61] has $N = 207$ nodes, $B = 213$ branches and $S = 1$ supply point, resulting in $\Lambda = 7$ open branches for each radial network, and in $\Psi = 151{,}641{,}612$ possible radial configurations. For larger networks the situation is even more extreme.

For example, in the urban distribution network used in [62] with $N = 535$ nodes, $B = 554$ branches and $S = 4$ supply points, there are $\Lambda = 23$ open branches for each radial network, and the number of possible radial configurations is as high as $\Psi = 7.197 \cdot 10^{32}$. With these numbers, it is clearly not feasible to determine the global optimum for this problem in a reasonable computation time. Thus, a number of deterministic and meta-heuristic methods have been proposed in the literature to calculate pseudo-optimal solutions with an acceptable computational burden. Reviewing the state of the art of these methods is outside the scope of this paper (some recent reviews may be found in [63] and in the introductory sections of [61]).

A further challenge of this problem is that the solutions have to satisfy two types of constraints:

1. All configurations created in the solution process have to be radial. This aspect is addressed by applying the close-open branch-exchange mechanism [64], in which each configuration change is applied by closing an open branch, detecting the closed loop formed in the network, then opening a branch in the loop to restore the radial structure.

2. Operational limits, referring to various aspects, such as maximum and minimum voltage magnitude limits at each node, maximum fault currents at each node, maximum current magnitude in each branch, and so forth [65]. This aspect is addressed by using a penalized objective function to ensure that any radial configuration corresponds to a numerical solution, discarding only the configurations leading to lack of convergence of the power flow calculations. In meta-heuristic solvers, a temporary solution worsening may be acceptable (if a probability-based check is passed) in order to open the search space to reach more solutions.

For a given radial configuration $X$, belonging to the set $\boldsymbol{X}$ of the radial configurations for which the power flow calculations is solvable, the total network losses $P_{tot}(X)$ are expressed in terms of the resistance $R_b$ and current magnitude $I_b$ of any branch $b \in \boldsymbol{B}$, where $\boldsymbol{B}$ is the set of network branches:

$$P_{tot}(X) = \sum_{b \in \boldsymbol{B}} R_b I_b^2 \tag{10}$$

Considering the set $\boldsymbol{V}$ of variables subject to constraint violation, the amount of violation $\Delta v_i$ and the penalty factor ($\rho_i > 0$ only if the corresponding violation exists) for $i \in \boldsymbol{V}$, the penalized objective function is:

$$f_p(X) = P_{tot}(X)[1 + \sum_{i \in V} \rho_i (\Delta v_i)^2] \quad (11)$$

The optimization problem becomes

$$\min_{X \in X} \{f_p(X)\} \quad (12)$$

Without loss of generality, the minimum losses problem addressed here is based on a single loading condition. More generally, with the diffusion of distributed resources in the distribution systems, the problem is formulated by considering the evolution in time of generation and demand patterns, assessing the losses on a given time interval, also with the possibility of determining an optimal set of network configurations for intra-day time periods [61]. The concepts introduced in this paper can be directly applied to the results of these more detailed problem formulations.

*5.3. Solution methods*

This paper considers the three meta-heuristic solvers Simulated Annealing (SA), Genetic Algorithms (GA), and Particle Swarm Optimization (PSO). These solvers have been widely used in the literature for the DSR problem [63]. For each solver, there is a set of parameters to be defined. In addition, these methods require random choices; repeatability of the results is ensured by fixing the *seed* for random number extraction to a given value $s_0$. The seed $s_0$ is the entry point to access the sequence of random numbers implemented in the specific function to extract random numbers from a uniform probability distribution in (0,1), and cannot be considered as a parameter of the solver.

The same *adaptive* stop criterion is used for the three methods, concluding the execution when the best solution does not change (or changes less than a predefined threshold) for a predefined number $N_s$ of successive iterations [65]. The other parameters are recalled below.

The SA method [41] is composed of a main cycle and an internal cycle. The main cycle does not require specific knowledge on the problem. It is driven by a control parameter $c$, whose initial value $c_0$ decays in the successive iterations (using $m$ as the iteration counter) with the progression $c_m = \alpha\, c_{m-1}$, with constant cooling rate $\alpha$ chosen in the range (0,1), until the adaptive stop criterion is satisfied. The initial value $c_0$ can be estimated by running the SA procedure with given values of the parameters until a predefined number of worse solutions $N_w$ has been reached, then calculating the average worsening $\overline{\Delta f_p}$ and imposing a percentage of acceptance $p_0$ for that worsening. Thereby, $c_0 = \overline{\Delta f_p}/\ln(1/p_0)$. At any iteration of the main cycle, an internal cycle is run. The internal cycle addresses the characteristics of the specific problem. Its implementation is simple: starting from the best configuration found so far, successive branch-exchanges are performed to maintain the network radial. In each branch exchange, the branches to close and open are chosen at random, and the resulting configuration is accepted if there is an improvement over the best solution found so far, or if the worsening is acceptable based on a probabilistic check. The internal cycle stops when the maximum number $M_A$ of configurations analyzed or the maximum number $M_C$ of configurations accepted are reached [65].

In the GA method [66], the information on the network configuration is coded in a string (called chromosome) containing $B$ binary values (called genes, set to 0 when the branch is open and to 1 when the branch is closed). A population of $C_{GA}$ strings is formed, the objective function is calculated for each string and its corresponding fitness is obtained by dividing the objective function values by the sum of the objective functions. In the iterative solution process, at each iteration the genetic operators *selection* (based on the fitness of each string, with a user-defined criterion), *crossover* (with crossover probability $p_c$) and *mutation* (with mutation probability $p_m$) are successively applied until the adaptive stop criterion is satisfied. While the selection operator is independent of the specific problem, the implementations of both crossover and mutation are critical, because the radial network configuration could be lost. As such, these operators have to be redefined to guarantee the generation of radial configurations. A trivial but time consuming solution would be to check if the network is radial after each application of crossover and mutation, rejecting the application of the operator for non-radial configurations. Effective solutions to guarantee that the configurations are radial have been proposed in [67] by implementing crossover only through a local improvement strategy (without using mutation), in [68] by using a node encoding based on Prufer numbers, and in [69] by implementing crossover and mutation by using the matroid theory. Further improvements for the mutation operator have been introduced in [70].

In the PSO method [40], the information is coded as in the GA. A population of $C_{PSO}$ particles is used. In the iterative solution process, the three terms forming the velocity of the particles (inertia, memory, and cooperation) are updated and the particle positions are determined for each particle. The objective function is calculated for each string, updating the local best and the global best each time the corresponding solution is improved. The inertia weight $w$ is progressively changed during the iterations from a maximum (initial) value $w_{init}$ to a minimum (final) value $w_{final}$ [71]. A list of possible strategies for varying the inertia weight $w$ is shown in [72]. In this paper, the strategy e1-PSO from [73] has been applied. In the PSO implementation, the knowledge on the specific problem is used to drive the creation of new radial configurations by using the information on the local best and the global best, as in [74]. The changes in the memory (or cooperation) term are applied only to a subset of branches chosen according to specific logical operations between the current configuration and the local (or global) best.

In summary, the sets of parameters considered are:
- for SA: $\boldsymbol{P}_{SA} = \{\alpha, N_w, p_0, M_A, M_C, N_s\}$;
- for GA: $\boldsymbol{P}_{GA} = \{C_{GA}, p_c, p_m, N_s\}$;
- for PSO: $\boldsymbol{P}_{PSO} = \{C_{PSO}, w_{init}, w_{final}, N_s\}$.

## 6. Application Examples

*6.1. Test network with known global optimum (Case G)*

Let us consider a test network used in various literature contributions, with *70 nodes* [75]. The characteristics of this network are summarized in Table 2. The total number of radial configurations is $\Psi = 407{,}924$ [61].

The initial network structure and loads are taken from [76], expressed in the reference system[2] using the rated voltage as the base voltage, and 10 kVA as the base power. The supply voltage is set to 1 p.u.

---

[2] This reference is adopted here in the absence of uniform definitions of base power and base voltages in the literature papers using this test network.

In the global optimum with these loading conditions the minimum total losses are 9.859 p.u.

*6.1.1. Comparison among the solutions obtained from different solvers*

For the execution of the solvers, $H_s = 100$ solutions are extracted from each solver, and $H = 100$ solutions are used to form the reference CDF. Furthermore, $N_s = 20$ is used in the stop criterion, and the seed for random number extraction is fixed ($s_0 = 1$) for all executions. The parameters of the three solvers are set up to leave only one or a few parameters variable in a given range of values (Table 3). The other parameters are defined with the following general choices:

- SA: the method illustrated in [65] is executed with $N_w = 10$, $p_0 = 0.5$, and constant values $M_A = 200$ and $M_C = 50$. The variable values of the cooling rate $\alpha$ are changed with regular steps.
- GA: the operators described in [67] and [70] are used for crossover and mutation, respectively. The mutation probability (applied to each gene) is set to $p_m = 0.001$. The variable parameters are the number of chromosomes $C_{GA}$ and the crossover probability $p_c$, each one with 10 values chosen at regular steps. The solver is run with all the combinations of the two parameters.
- PSO: the method described in [74] has been used. The variable parameters are the number of particles $C_{PSO}$ and the initial value of the inertia weight $w_{init}$ (with the final value of the inertia weight fixed to $w_{final} = 0.4$ [71] and the variation strategy taken from [73]), each of which with 10 values chosen at regular steps. The solver is run with all the combinations of the two parameters.

The set of values used in this example aim to generate various solutions and calculate the indicator $OPISD_G^{(H)}$, without any attempt to find the optimal set of parameters for each method.

Fig. 8 shows the reference CDF and the test CDFs of the results obtained from the methods to be compared. Table 4 reports the performance indicators. The solutions provided by SA and PSO cover 100% of the reference CDF, reaching the global optimum. These methods are then viable to be tested on large-scale cases. The example shown for the GA solutions suggests that the setting used for GA is inappropriate to be used, so this GA is not suggested to be used any further.

*6.2. Large real network with unknown global optimum (Case R)*

The 207-node real distribution network mentioned in Section 4.2 is used for the calculations. The characteristics of this network are summarized in Table 5. The base power is 1 MVA, and the base voltage is equal to the rated voltage of the network (10.5 kV). The global optimum is unknown. The methods SA, GA and PSO have been tested with the same parameters indicated in Table 3. The number of solutions extracted for each solver is $H_s = H = 100$.

*6.2.1. Comparison among the solutions obtained from different solvers*

Fig. 9 shows the reference CDF and test CDFs of the results obtained from the three methods. Table 6 reports the performance indicators. In the loading conditions considered, the SA solutions are the best ones in the *OPISD* ranking, followed by PSO and GA. Fig. 9 shows that the SA results are not uniformly located at the left-hand side of the CDFs referring to PSO and GA. The SA solution with the highest objective function is worse than the PSO solution with the highest objective function.

In the case studied, the *PERC* ranking provides the same results, i.e., the performance of SA and PSO are significantly better than the performance of GA. However, Table 6 indicates that the PSO results are closer to the ones of GA rather than SA, because the comparison is just based on the number (and percentage) of occurrences of the best solution, without taking into account the distribution of all the other solutions.

*6.2.2. Variation of the SA solutions for different seeds for random number extraction*

As a further illustrative example, the SA solver is taken, all its parameters are set up to given values, and only the initial seed for random number extraction is changed. In these conditions, the solutions generally change in different executions, especially when the solver is run on a large-scale system. In order to show an example of parametric analysis conducted according with the principles indicated in this paper, four cooling rate values are considered, namely, $\alpha = \{0.2, 0.5, 0.7, 0.95\}$, leaving all the other parameters constant and equal to the values used in Section 5.1.1. These values are intentionally very different, also going down to 0.5 and 0.2, well-known from the general theory of SA as poorly appropriate, since they may provide excessively fast decay of the control parameter in the external cycle. The rationale of this choice is to test whether the proposed indicators correctly represent the appropriateness of the parameter setting. For each cooling rate, $H = 120$ solutions have been found, building the corresponding test CDFs. Fig. 10 shows the reference CDF and the test CDFs for the four cooling rate values, whereas Table 7 shows the corresponding performance indicators defined in the previous sections. The results confirm the suitability of the cooling rate 0.95, while, with the other cooling rates, the SA reaches the pseudo-optimal solution for a lower number of times and falls into worse solutions in most cases. As expected, *OPISD* ranks the CDFs from the one having the higher cooling rate to the one with the lower cooling rate.

## 7. Concluding Remarks

This paper has discussed a number of aspects to compare different solvers for heuristic optimization. A conceptual scheme has been drawn to represent the process used for claiming that a new "best solver" has been found. From this scheme, it may be seen that the mechanism of creation and testing of new heuristics cannot be formally stopped. However, the definition of comparison metrics more robust than simple but weak indicators (such as mean value, median, standard deviation, best value, worst value, as well as the *PERC* indicator shown in this paper) may help reducing the proposal of further solvers without a clear superior performance with respect to other solvers used for the same problem. Establishing a sound comparison requires the definition of a benchmark solver for each problem, whose outcomes may be expressed as the CDF composed of *H* solutions. On these bases, the comparisons with the results obtained by other solvers may be carried out by calculating the *OPISD* indicators corresponding to the benchmark solver and the proposed solvers, setting up the CDF of the solutions given by the top ranked solver as the (unchanged or new) benchmark. More generally, testing the solvers on multiple problems is important for a broader comparison among the solvers, with the aim of gaining some kind of confidence that the solvers could perform well on unseen problems. However, for the solvers applied to electrical energy system problems, it has to be considered that the same solver could be more suitable to be applied to a problem rather than another, for example because the constraints can be modeled in an easier way for that problem, while for another problem the modeling of the constraints could be more complicate (and this complication does not occur for all the solvers).

On the practical point of view, the proliferation of "new" best solvers could be limited by introducing both more robust indicators to compare the solvers, and suitable benchmarks validated by the power and energy system community. An example is the $OPISD_G$ indicator used to carry out a *pre-testing* of a proposed heuristic on cases with known global optimum, to verify that 100% of the solutions are found ($OPISD_G = 1$), before using that heuristic for further studies. The examples shown in this paper have confirmed that some solvers that exhibit good properties on large networks have no problem in finding the global optimum in 100% of the executions on networks for which the global optimum is known. In addition, the solvers to be compared have to be chosen in an effective way, to avoid comparisons among solvers developed in a detailed way and basic versions of other solvers implemented with limited effort to incorporate the characteristics and constraints of the problem.

In summary, the concepts presented in this paper, supported by the application cases, suggest the following remarks:

- In order to filter out the contributions based on poorly meaningful definitions of "better solution", the papers that show comparisons only based on classical simple indicators do not provide effective advancements of the state of the art.

- Robust testing of the results obtained by the solver has to be carried out by using effective indicators with statistical meaning, such as the ones based on the first order stochastic dominance presented in this paper.

- For a problem with known global optimum, any solver (with its solution method and parameter settings) that generates the globally optimal solution in less than 100% of the executions should be considered inadequate to be used on the same problem for systems with larger size. Likewise, contributions presenting only the testing of a heuristic algorithm on a problem with known global optimum have an excessively limited scope and do not provide sufficient insights on the viability of application of the heuristic.

- For a problem with unknown global optimum, the ranking of the solutions by using *OPISD* may be useful to compare different solvers, or the same solver with different parameters, trying to create a solid benchmark. However, it has to be noted that, in heuristic methods depending on random extractions, even the change of the seed for random number extractions could generate different outcomes. As such, the results obtained on any benchmark that may be constructed cannot be considered as an absolute reference.

- The proposed indicators may also be useful to detect improvements in the construction of a solver, to avoid another source of proliferation of publications that try and propose variants of the same solver (or hybridizations of different solvers) as the new "best" solvers just on the basis of the best/mean/median/standard deviation value obtained from some tests.

# Tables

Table 1.
Number of articles from Science Direct (at 25 August 2018) that mention or apply a heuristic solver on a given problem.

| Heuristic name | Problem | | | | | |
|---|---|---|---|---|---|---|
| | DSR | ED | LF | MS | OPF | PSP |
| Ant colony optimization (ACO) [15] | 225 | 458 | 226 | 129 | 293 | 91 |
| Artificial bee colony (ABC) [16] | 86 | 336 | 111 | 24 | 247 | 55 |
| Artificial immune system (AIS) [17] | 38 | 64 | 44 | 33 | 84 | 17 |
| Bacterial foraging (BF) [18] | 75 | 210 | 62 | 9 | 136 | 24 |
| Bat algorithm (BA) [19] | 32 | 71 | 31 | 8 | 58 | 10 |
| Big-bang big-crunch (BBBC) [20] | 32 | 26 | 9 | 2 | 25 | 8 |
| Biogeography based optimization (BBO) [21] | 21 | 171 | 22 | 13 | 128 | 31 |
| Charged system search (CSS) [22] | 3 | 28 | 6 | 2 | 17 | 4 |
| Cuckoo search algorithm (CSA) [23] | 45 | 140 | 73 | 16 | 97 | 21 |
| Differential evolution (DE) [24] | 122 | 702 | 217 | 70 | 456 | 115 |
| Evolutionary algorithms (EA) [25] | 221 | 838 | 445 | 171 | 534 | 157 |
| Evolution strategies (ES) [26] | 32 | 136 | 48 | 31 | 61 | 19 |
| Firefly algorithm (FA) [27] | 48 | 163 | 96 | 18 | 109 | 31 |
| Flower pollination algorithm (FPA) [28] | 8 | 40 | 13 | 2 | 31 | 6 |
| Fruit fly optimization (FFO) [29] | 2 | 22 | 49 | 4 | 4 | 3 |
| Genetic algorithms (GA) [30] | 611 | 1629 | 1092 | 696 | 1164 | 407 |
| Grey wolf optimization (GWO) [31] | 10 | 66 | 14 | 5 | 49 | 8 |
| Gravitational search algorithm (GSA) [32] | 42 | 212 | 78 | 14 | 191 | 37 |
| Group search optimization (GSO) [33] | 15 | 96 | 16 | 4 | 59 | 7 |
| Harmony search algorithm (HSA) [34] | 106 | 340 | 82 | 31 | 237 | 50 |
| Imperialist competitive algorithm (ICA) [35] | 7 | 29 | 4 | 2 | 18 | 5 |
| Invasive weed optimization (IWO) [36] | 7 | 38 | 4 | 3 | 33 | 6 |
| Krill herd algorithm (KHA) [37] | 10 | 42 | 14 | 3 | 49 | 3 |
| Memetic algorithms (MA) [38] | 22 | 78 | 32 | 25 | 27 | 11 |
| Moth-flame optimization (MFO) [39] | 0 | 13 | 4 | 0 | 9 | 2 |
| Particle swarm optimization (PSO) [40] | 346 | 1281 | 706 | 208 | 895 | 259 |
| Simulated annealing (SA) [41] | 255 | 650 | 333 | 263 | 410 | 149 |
| Scatter search (SS) [42] | 5 | 24 | 4 | 12 | 10 | 5 |
| Seeker optimization algorithm (SOA) [43] | 7 | 48 | 11 | 2 | 33 | 11 |
| Shuffled frog leaping algorithm (SFLA) [44] | 50 | 91 | 26 | 11 | 102 | 31 |
| Social spider algorithm (SSA) [45] | 2 | 12 | 4 | 0 | 6 | 0 |
| Symbiotic organisms search algorithm (SOS) [46] | 5 | 13 | 5 | 1 | 13 | 2 |
| Teaching-learning-based optimization (TLBO) [47] | 44 | 167 | 31 | 9 | 158 | 30 |
| Tabu search (TS) [48]* | 198 | 398 | 139 | 184 | 284 | 91 |
| Whale optimization algorithm (WOA) [49] | 2 | 9 | 5 | 2 | 16 | 2 |

*The TS has been inserted as a matter of comparison, even if it is not a probability-based method

Table 2
Test Network Data

| | |
|---|---|
| rated voltage [kV] | 12.66 |
| number of nodes $N$ | 70 |
| number of branches $B$ | 74 |
| number of supply points $S$ | 1 |
| number of open branches $\Lambda$ | 5 |
| number of radial configurations $\Psi$ | 407,924 |

Table 3
Parameters Used for Computing the Set of Solutions

| solver | parameter | range of values | step |
|---|---|---|---|
| SA | $\alpha$ | 0.900÷0.999 | 0.001 |
| GA | $C_{GA}$ | 100÷190 | 10 |
|  | $p_c$ | 0.35÷0.44 | 0.1 |
| PSO | $C_{PSO}$ | 100÷190 | 10 |
|  | $w_{init}$ | 0.81÷0.90 | 0.01 |

Table 4
Performance Indicators Calculated for the Test System

| network | solver | $PERC_G^{(H)}$ | $A_{s,G}^{(H)}$ | $OPISD_G^{(H)}$ |
|---|---|---|---|---|
| 70 nodes | SA | 100% | 0 | 1 |
|  | PSO | 100% | 0 | 1 |
|  | GA | 94% | 0.0180 | 0.9823 |

Table 5
Real Network Data

| | |
|---|---|
| rated voltage [kV] | 10.5 |
| number of nodes $N$ | 207 |
| number of branches $B$ | 213 |
| number of supply points $S$ | 1 |
| number of open branches $\Lambda$ | 7 |
| number of radial configurations $\Psi$ | $1.516 \cdot 10^5$ |

Table 6
Performance Indicators Calculated for the Real Network

| network | solver | $PERC_G^{(H)}$ | $A_{s,G}^{(H)}$ | $OPISD_G^{(H)}$ |
|---|---|---|---|---|
| 207 nodes | SA | 59% | 0.0165 | 0.9837 |
|  | PSO | 20% | 0.0464 | 0.9556 |
|  | GA | 0% | 1.9533 | 0.3386 |

Table 7
Performance Indicators Calculated with the SA Method (Real Network)

| network | cooling rate $\alpha$ | $PERC_G^{(H)}$ | $A_{s,G}^{(H)}$ | $OPISD_G^{(H)}$ |
|---|---|---|---|---|
| 207 nodes | 0.2 | 57% | 0.0024 | 0.9976 |
|  | 0.5 | 53% | 0.0019 | 0.9981 |
|  | 0.9 | 69% | 0.0011 | 0.9989 |

# Figures

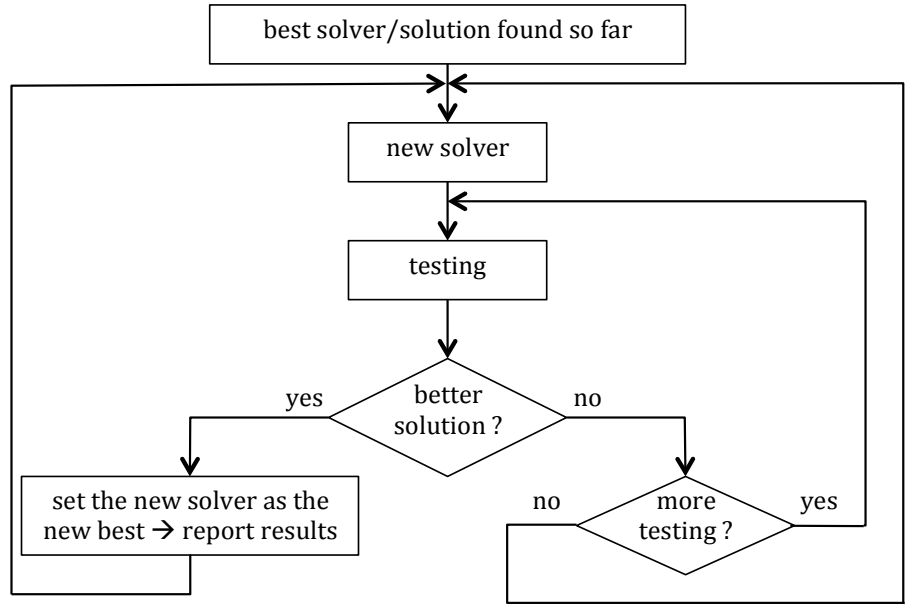

Fig. 1. Comparing optimization solvers on the basis of their solutions – A *perpetual motion* conceptual scheme.

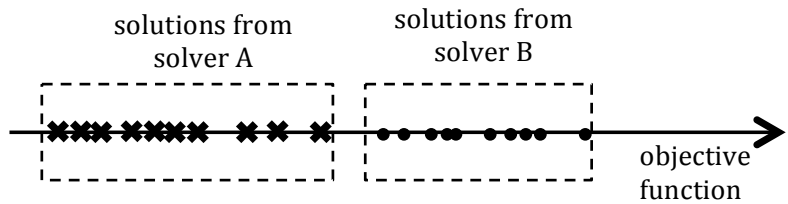

Fig. 2. Deterministic dominance. For the objective function minimization, the solutions from solver A dominate the solutions from solver B on a deterministic way.

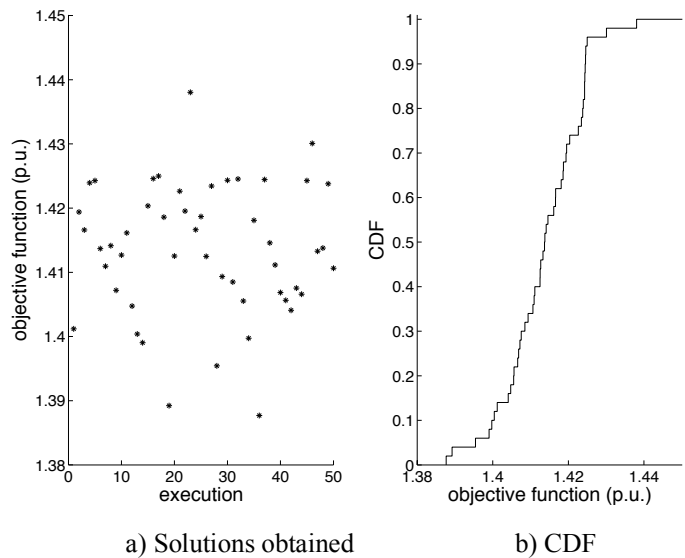

a) Solutions obtained  b) CDF

Fig. 3. Construction of the CDF.

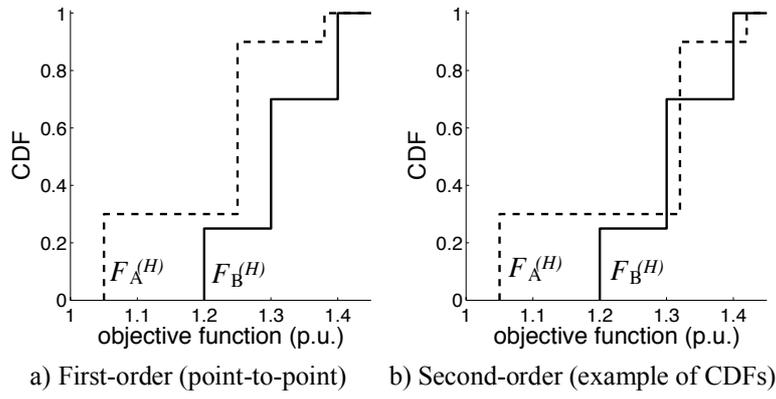

a) First-order (point-to-point)   b) Second-order (example of CDFs)

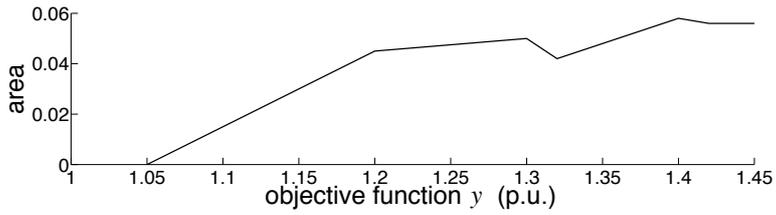

c) Positive areas $\int_{z=0}^{y}\left(F_A^{(H)}(z)-F_B^{(H)}(z)\right)$ satisfying the second-order stochastic dominance conditions

Fig. 4. Stochastic dominance.

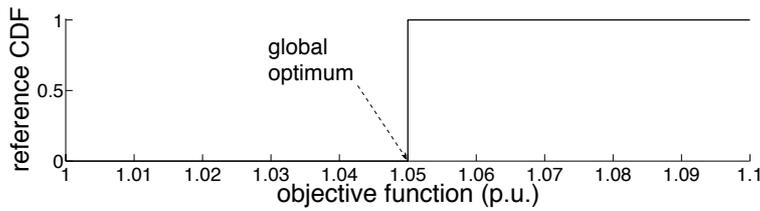

Fig. 5. Reference CDF for *Case G*.

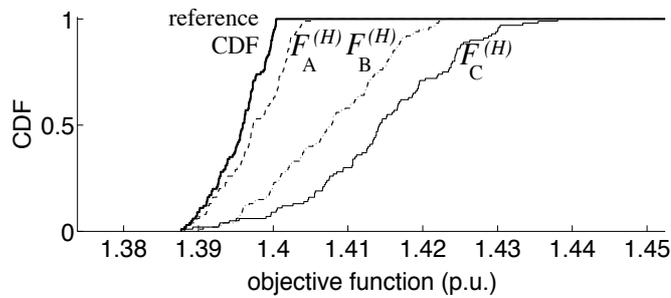

Fig. 6. Reference CDF for *Case R*.

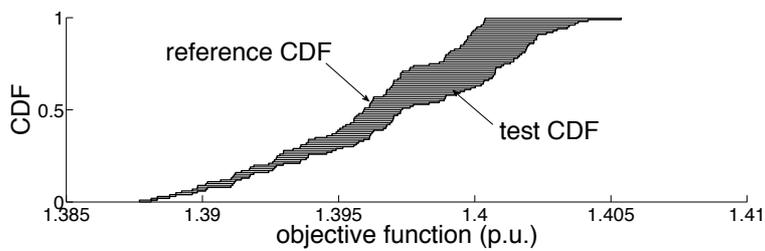

Fig. 7. Representation of the areas contributing to the *OPISD* definition.

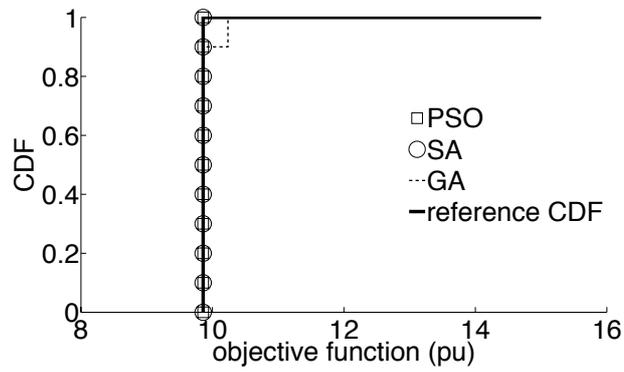

Fig. 8. Reference CDF and test CDFs for the methods run on the 70-node test system.

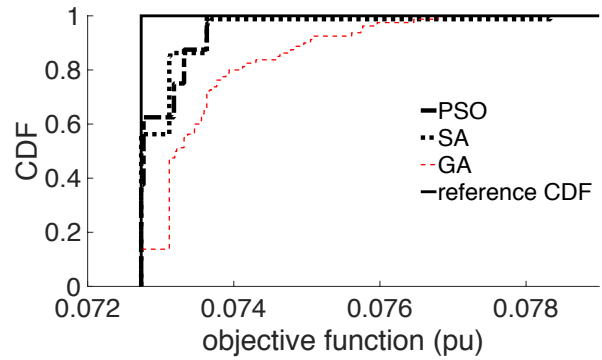

Fig. 9. Reference CDF and test CDFs for the three methods (207-node network). The internal zoom refers to SA and PSO close to the lowest objective function values.

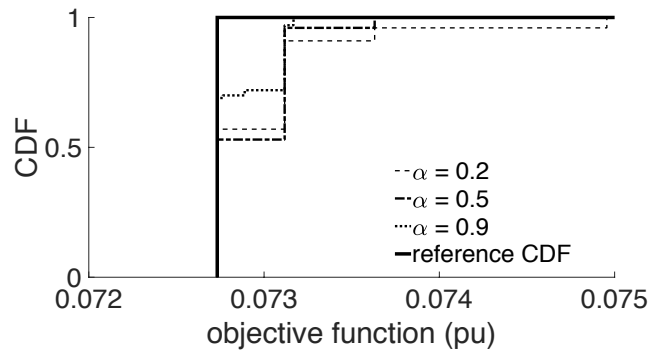

Fig. 10. Reference CDF and test CDFs for the three methods (207-node network, SA with different cooling rate $\alpha$ and 120 executions for each cooling rate with different initial seed for random number extraction).

**Highlights**

- The search of the best solver for heuristic methods is not a well-posed problem.
- A scheme that explains the unlimited introduction of "best solvers" is shown.
- Two first-order stochastic dominance-based metrics are proposed.
- These metrics allow more significant comparison among optimization solutions.
- Using these metrics can reduce the uncontrolled proliferation of new "best solvers".